\documentclass[twocolumn,pra,superscriptaddress]{revtex4-2}
\usepackage[english]{babel}
\usepackage[utf8]{inputenc}
\usepackage[colorinlistoftodos, color=green!40, prependcaption]{todonotes}
\usepackage{color}
\usepackage{float}
\usepackage{amsthm}
\usepackage{subcaption}
\usepackage{makecell}
\usepackage{placeins}
\usepackage{mathtools}
\usepackage{amsmath,amsfonts,amssymb}
\usepackage{physics}
\usepackage{xcolor}
\usepackage{graphicx}
\usepackage{tabularx}
\usepackage{array}
\usepackage{booktabs}
\usepackage{unicode-math}
\usepackage{lastpage}
\usepackage{circuitikz} % for logical gates
\usepackage{qcircuit} % quantum circuits
\usepackage{url}
\usepackage{soul}
\usepackage{bm}% bold math
\usepackage{mathbbol}
\usepackage{rotating} % in preamble
\usepackage{caption}

\definecolor{thepurple}{rgb}{0.5882352941176471, 0.45098039215686275, 0.6509803921568628}
\definecolor{citecolor}{rgb}{0.0, 0.2, 0.45}
\definecolor{linkcolor}{rgb}{0.55, 0.0, 0.35}
\definecolor{urlcolor}{rgb}{0.0, 0.2, 0.55}

\newcolumntype{Z}{>{\centering\let\newline\\arraybackslash\hspace{0pt}}X}
\begin{document}
\title{Quantum-Inspired Self-Attention in a Large Language Model}

\author{Nikita Kuznetsov}
\affiliation{Higher School of Economics, National Research University, St. Petersburg, Russian Federation}
\affiliation{JV Quantum LLC, State Atomic Energy Corporation Rosatom, Moscow, Russian Federation}
\email{nvkuznetsov\_4@edu.hse.ru}

\author{Niyaz Ismagilov}
\affiliation{JV Quantum LLC, State Atomic Energy Corporation Rosatom, Moscow, Russian Federation}

\author{Ernesto Campos}
\altaffiliation{Former affiliation}
\affiliation{Skolkovo Institute of Science and Technology, Moscow, Russian Federation}

\begin{abstract}

Recent advances in Natural Language Processing have been predominantly driven by transformer-based architectures, which rely heavily on self-attention mechanisms to model relationships between tokens in a sequence. Similarly, the field of Quantum Natural Language Processing, which seeks to leverage quantum principles to address challenges in language understanding and generation tasks, has seen the recent development of quantum self-attention mechanisms. 
We propose a classical quantum-inspired self-attention (QISA) mechanism  and integrate it into the full autoregressive language modeling pipeline of GPT-1. To the best of our knowledge, this is the first integration of this kind, as previous quantum self-attention mechanisms have been primarily tested on text classification. In our experiments, QISA achieves better performance when compared to standard self-attention on the metrics character error rate ($15.5\times$ better), word error rate ($4.7 \times $) and cross-entropy loss ($13 \times$). This is achieved while only requiring a $ 2.6\times$ longer inference time. 

\end{abstract}

\maketitle

\section{Introduction}

Transformer-based language models have fundamentally reshaped modern Natural Language Processing (NLP), enabling major advances in tasks ranging from language modeling to reasoning and generation \cite{vaswani2017attention, devlin2018bert}. Central to this progress is the self-attention mechanism, which provides a flexible way to model long-range dependencies and contextual interactions between tokens. However, the continued scaling of attention-based architectures has led to rapidly increasing computational and memory requirements, motivating the search for alternative modeling paradigms that can offer improved efficiency and representational capacity \cite{preskill2018quantum, cerezo2021variational}.
Quantum computing presents a promising avenue for enhancing classical NLP models through its unique computational properties \cite{huang2021near, lloyd2018quantum}. Specifically, quantum systems can efficiently represent and process high-dimensional data through quantum superposition and entanglement, potentially enabling more compact and expressive models than their classical counterparts \cite{biamonte2017quantum, schuld2019quantum}. The field of Quantum Natural Language Processing (QNLP) has emerged at this intersection, seeking to leverage quantum principles to address challenges in language understanding and generation tasks \cite{zeng2016quantum, coecke2020foundations}.

Analogous to NLP, much of the recent advances in the field of QNLP have come from the development of quantum self-attention (QSA) structures. These QSAs \cite{li2024quantum,zheng2023design} follow the variational model of quantum computing, where a parameterized quantum circuit is iteratively optimized to minimize a given cost function. These implementations of QSA have mostly focused on text classification tasks where they demonstrate advantage in terms of accuracy / F1 score, while offering a reduction on the number of trainable parameters.
Among these approaches, the quantum self-attention neural network (QSANN) introduced in \cite{li2024quantum} stands out for having one of the lowest sampling complexities, while still claiming a performance uplift over the classical models. 
Despite this advantage, its structure limits the degree in which QSANN can be parallelized, a core strength of the transformer architecture that led to its dominance over other models used in NLP like the recurrent neural networks \cite{schmidt2019recurrentneuralnetworksrnns},
and long-short term memory \cite{vennerod2021longshorttermmemoryrnn}.
Taking cues from the framework of QSANN, we propose a classical quantum-inspired self-attention mechanism (QISA). Specifically, the proposed mechanism replaces the standard value layer in the classical self-attention (CSA) \cite{vaswani2017attention} by operations inspired by quantum computing. QISA inherits the strengths of both classical and quantum self-attention; it can be classicaly parallelized while having a performance that broadly exceeds that of QSANN. Moreover, we propose a variant, called QISA-A, which can be implemented in quantum devices and displays similar performance. Although this variant is slower than QISA when classically simulated, its lower parameter count may offer an advantage when implemented on future error-corrected quantum devices.   

For benchmarking, we integrate QISA, the classical simulation of QISA-A, three QSANN variants, and CSA, into the full autoregressive language modeling pipeline of GPT-1 \cite{vaswani2017attention}. To the best of our knowledge, this integration of QSAs is the first of its kind. The experiments show that QISA performs on a par with the QSAs and outperforms CSA in the metrics character error rate (CER),  word error rate (WER) and cross-entropy loss, with improvements of 15.5$\times$, $4.7\times$ and $13\times$ respectively. The timed duration of each inference step shows that our implementation of QISA has a reasonable $2.6\times$ longer duration compared to CSA for several cases, making it an attractive alternative for current language models.    

The paper is structured as follows: Section II reviews the classical multi‑head self‑attention mechanism and the variational quantum algorithmic framework; In Section III we present the details of the QISA and QISA-A models; Section IV details the integration of QISA and other QSAs into GPT‑1, and discusses the experimental results; and Section V shows conclusions.

\section{Methods}

\subsection{Classical multi-head self-attention structure}

The multi-head self-attention (MHSA) mechanism \cite{vaswani2017attention} allows the model to jointly attend to information from different representation subspaces at multiple positions, improving expressiveness and enabling the network to capture various types of relationships between tokens in parallel.
It is computed as follows:

\begin{enumerate}
    \item Input: sequence of embeddings $X = [x_1~ x_2~ \dots~ x_l] \in \mathbb{R}^{l\times m}$,
where \( x_i \in \mathbb{R}^m \).

\item  Linear Transformations:
 Given the number of heads $H$ and head size $h=m/H$, project the input embeddings into three sets of matrices resulting in query, key and value matrices: $Q^{(j)} = XW_Q^{(j)}, \quad K^{(j)} = XW_K^{(j)}, \quad V^{(j)} = XW_V^{(j)}$, where \( W_Q^{(j)}, W_K^{(j)}, W_V^{(j)} \in \mathbb{R}^{m \times h} \), and $j \in [1,H]$.

\item Attention Scores:
Compute the scaled dot-product between the query and key matrices with normalization: 
\begin{equation}
    A^{(j)} = \mathrm{Softmax}\left(\frac{Q^{(j)}\,K^{(j)T}}{\sqrt{h}}\right)\in\mathbb{R}^{l\times l}.
    \label{eq:attention}
\end{equation}
where $\mathrm{Softmax}$ is applied row-wise. For a vector $x \in \mathbb{R}^l$, it is defined as
\begin{equation}
    \mathrm{Softmax}(x) = \frac{e^{x}}{\sum_{k=1}^{l} e^{x_k}}.
\end{equation}
\item Output:
Compute the weighted sums $\mathrm{head}^{(j)} = A^{(j)}\,V^{(j)}\in\mathbb{R}^{l\times h}$.
% \todo[inline]{EC: this dimension does not match, please check, I think it should be $l \times h$}
% \todo[inline]{NK: Yes, the typo was in incorrect dimensions for X, now everything should be ok}

Then concatenate and project: $\mathrm{MultiHead}(X) = \bigl[\mathrm{head}^{(1)},\dots,\mathrm{head}^{(H)}\bigr]\,W_O\;\in\mathbb{R}^{l\times m}$,  where $W_O \in \mathbb{R}^{m\times m}$.

\end{enumerate}

% \todo[inline]{the self atention mechanis mor some of its components copuld be replaced by quantum as shown ...}

\subsection{Variational quantum algorithms}

Variational quantum algorithms (VQAs) comprise a class of hybrid quantum–classical methods tailored to the capabilities of Noisy Intermediate–Scale Quantum (NISQ) devices \cite{cerezo2021variational}. A VQA employs a parameterized quantum circuit (ansatz) \(U(\theta)\) acting on \(n\) qubits to prepare a trial state
\begin{equation}
    |\psi(\theta)\rangle = U(\theta)\,|0\rangle^{\otimes n}, \label{eq:quantum_evolution}
\end{equation}
where \(\theta = (\theta_1,\theta_2,\dots,\theta_p)\) denotes the vector of tunable gate parameters. Given a set of Hermitian observables \(\{O_k\}\), the quantum processor estimates expectation values
\begin{equation}
   \langle O_k\rangle_\theta = \langle\psi(\theta)\,|\,O_k\,|\,\psi(\theta)\rangle,
\end{equation}
which are subsequently combined via a classical cost function
\begin{equation}
    C(\theta) = f\bigl(\langle O_1\rangle_\theta ,\langle O_2\rangle_\theta ,\dots\bigr),
\end{equation}
designed to quantify task performance (e.g.\ ground‐state energy, classification loss, or feature‐map fidelity). A classical optimizer iteratively updates the parameters to minimize \(C(\theta)\). 

% based methods usually obtained by the parameter‐shift rule \cite{Wierichs_2022} - requiring two expectation values per parameter per observable - or uses gradient‐free methods such as COBYLA \cite{Pellow_Jarman_2021} or SPSA \cite{resch2021introductorytutorialspsaquantum}.

Ansatz architectures are chosen to balance expressivity against circuit depth and qubit count, e.g. hardware‐efficient ansatzes \cite{Leone_2024} interleave parameterized single‐qubit rotations with entangling gates in a low‐depth topology, while problem‐inspired circuits embed data structure directly into the gate sequence \cite{lee2018generalized, farhi2014quantum, wecker2015progress}. 

% The VQA paradigm is particularly suited for quantum-inspired self‐attention, as it captures high‐order correlations via entanglement and measurement while remaining within NISQ‐era resource constraints.
In QISA, we take inspiration from variational circuits and  quantum expectation values to compute the value vectors, thereby inheriting the expressivity of quantum feature maps without exceeding classical simulation limits.

\section{Quantum-inspired self-attention structure}
We propose the use of quantum-inspired value layers $\tilde{V}^{(j)}$ in the classical self-attention structure. We take inspiration from the value layer in QSANN \cite{li2024quantum} (see Appendix \ref{App:QSANNv0}). This choice comes from our numerical experiments that identify the value layer of QSANN as the main contributor to its improvement over CSA (as it is later shown in Section \ref{sec:experiments}).

In our approach, each input token is a normalized classical vector $\ket{x_i} = [x_{i1}, x_{i2}, \cdots , x_{im}]$. 
% This is analogous to a quantum state with amplitudes proportional to its elements,
% \begin{equation}
%     \ket{x_i} = c\sum_{j=1}^m x_{ij}\ket{j}, \label{eq:amplitude_encoding}
% \end{equation}
% where $\ket{j}$ are computational basis states.

The elements of $\tilde{V}^{(j)}$ corresponding to each token are calculated independently of each other as
\begin{equation}
    v_i^{(j)} := [\langle P_1\rangle_i^{(j)}~ \langle P_2\rangle_i^{(j)}~ \cdots~ \langle P_h\rangle_i^{(j)}]
\end{equation}
\noindent where
\begin{equation}
    \langle P_k\rangle_i^{(j)} = \bra{x_i} \tilde{W}_V^{(j)\intercal} P_k  \tilde{W}_V^{(j)} \ket{x_i},\label{eq:P_k}
\end{equation}
 $\tilde{W}_V^{(j)}$ is a trainable linear map in $\mathbb{R^{m \times m}}$, and $P_k \in \{I, X, Y, Z\}^{\otimes n}$ is a Pauli string.
Finally, $\tilde{V}^{(j)}$ is given by the concatenation of $\{v_i^{(j)}\}_{i=1}^l$,
\begin{equation}
    \tilde{V}^{(j)}=[v_1^{(j)}~ v_2^{(j)}~\cdots ~ v_l^{(j)}]
\end{equation}
The calculations of each query and key layers remain the same as in the classical case. As such, in our approach the operators $W_Q^{(j)}, W_K^{(j)}$, and  $\tilde {W}_V^{(j)}$ are token independent, in contrast to QSANN which requires the training of an ansatz circuit per token (see Appendix \ref{App:QSANNv0}). 

The linear map $\tilde{W}_K^{(j)}$ in \eqref{eq:P_k} is analogous to the ansatz circuit $U(\theta)$ in \eqref{eq:quantum_evolution}. We opt for using a classical linear map to avoid the overhead of calculating the matrix form of $U(\theta)$. This is particularly relevant during training, where $U(\theta)$ needs to be recalculated for different parameter values at each iteration. This is clearly only true for simulation, since an implementation in quantum hardware would natively perform the unitary operations. We call QISA-A the model variant that exchanges $\tilde{W}_K^{(j)}$ for a quantum ansatz $U(\theta)$.

In our numerical experiments, both QISA and QISA-A obtained near identical performance across multiple metrics (as shown in Section \ref{sec:experiments}). It is yet to be determined whether an implementation on error-corrected quantum hardware \cite{gottesman2022opportunities, roffe2019quantum} would offer an advantage in optiimization and/or inference speed over the fully classical QISA. Although inference and optimization are likely to benefit from techniques such as classical shadows \cite{huang2020predicting, koh2022classical}, overheads could appear in the form of: training limiting effects \cite{larocca2025barren, bittel2021training, anschuetz2022quantum, campos2021abrupt}, the use of the parameter shift rule \cite{Wierichs_2022}, and amplitude  encoding for input tokens \cite{gonzalez2024efficient}.

\section{Implementation Details And Experiments}
\label{sec:experiments}

\begin{figure}[ht!]
    \includegraphics[width=0.5\textwidth]{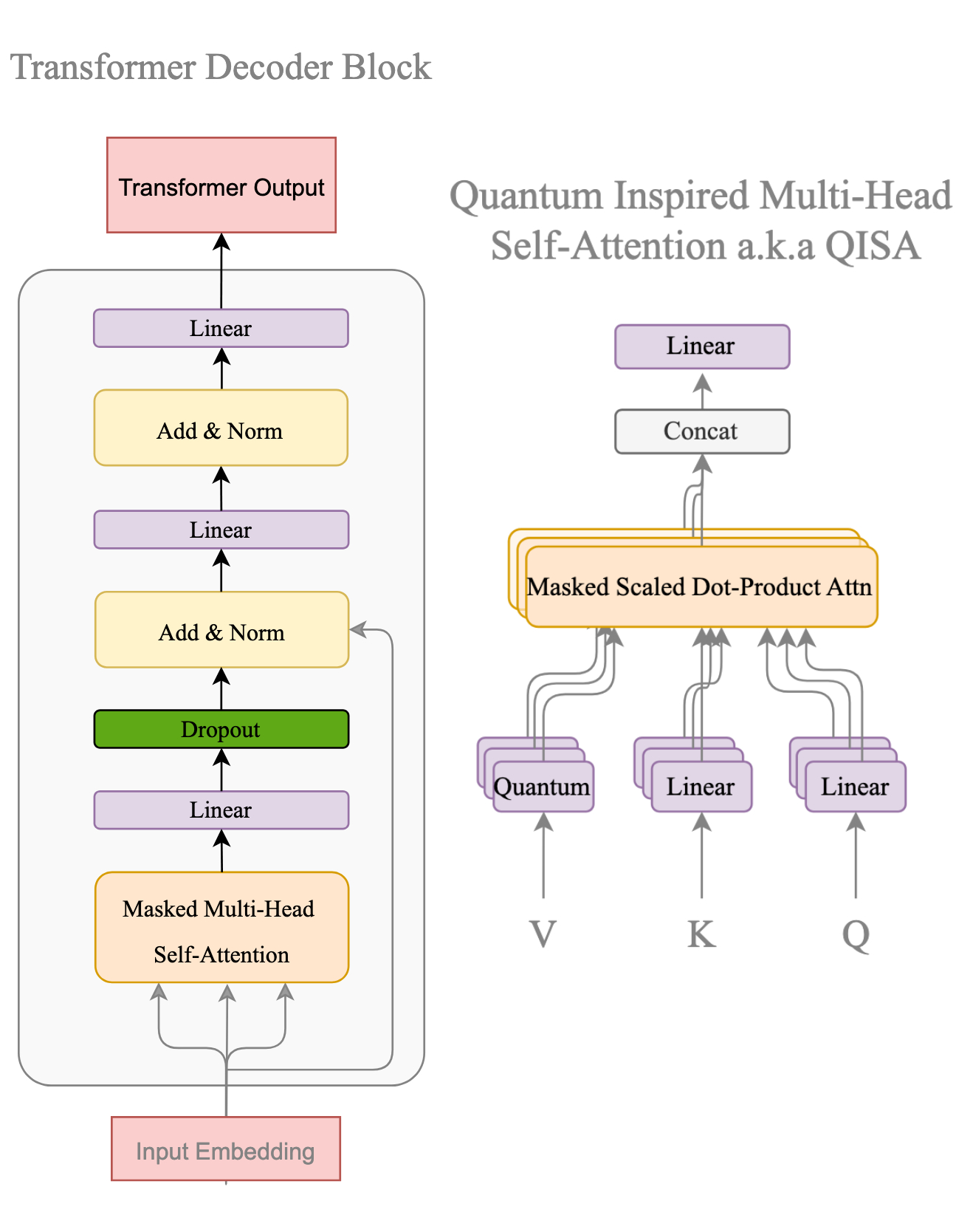}
    \caption{Left: Standard transformer block structure used in language models. Right: Modified multi-head self-attention where the standard value layer is replaced with a quantum-inspired one.}
    \centering
    \label{fig:qgpt}
\end{figure}

Our model retains the original GPT-1 structure \cite{vaswani2017attention} but with a key modification: we replace the standard attention heads with one of six distinct classical or quantum (simulated) variants, forming the basis of our experimental comparison. A graphical high-level diagram of a transformer block is shown in Figure \ref{fig:qgpt}.

% Let us discuss the details of the implementation of the QISA in GPT-1. Here, we use amplitude encoding as in \eqref{eq:amplitude_encoding} due to its ease of implementation in simulation.
To preserve the autoregressive property, we apply the standard lower-triangular mask. Let $A\in\mathbb{R}^{l\times l}$ be the pre-softmax attention logits \eqref{eq:attention}; then
\[
A_{ij} = -\infty\quad\text{for }j>i,
\]
so that after $\mathrm{Softmax}(A)$ each token $i$ only attends to positions $\le i$.  

\begin{figure}[h]
    \centering
    \centering
    \includegraphics[width=0.9\linewidth]{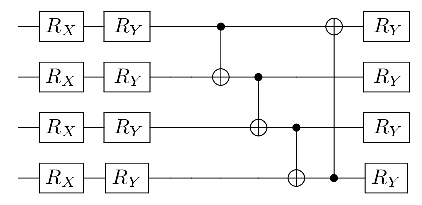}
    \caption{One layer of the hardware efficient ansatz.}
    \label{fig:hea_ansatz}
\end{figure}

For our experiments, GPT-1 is reimplemented using PyTorch, and quantum self-attention mechanisms use the TorchQuantum framework \cite{hanruiwang2022quantumnas}. As a dataset, we use Shakespeare's texts \cite{adarshpathak_shakespeare_text_2026}, with 20$\%$ reserved for testing, and a character-level tokenizer.
We consider the following three configurations: embedding sizes 4 and 16 with 1 head, and embedding size 16 with 4 heads. All these configurations have 6 transformer layers and a context size of 16 tokens.
The tests are conducted for six different self-attention models: CSA, QISA, QISA-A, and three variants of QSANN. In addition to the original QSANN, we introduce two variants: (i) QSANNv1, which uses fewer parameters (see Appendix~\ref{app:QSANNv1}), and (ii) QSANNv2, which uses a more expressive structure for $Q$/$K$ (see Appendix~\ref{app:QSANNv2}).
All quantum models (QISA-A, QSANN, QSANNv1, and QSANNv2) were tested with 1 to 3 layers of the hardware efficient ansatz (HEA), illustrated in Figure \ref{fig:hea_ansatz}.

\begin{figure}[h]
    \centering

    \begin{subfigure}{\linewidth}
        \centering
        \textbf{Embedding size = 4, Number of heads = 1}\par\vspace{2mm}
        \includegraphics[width=1\linewidth]{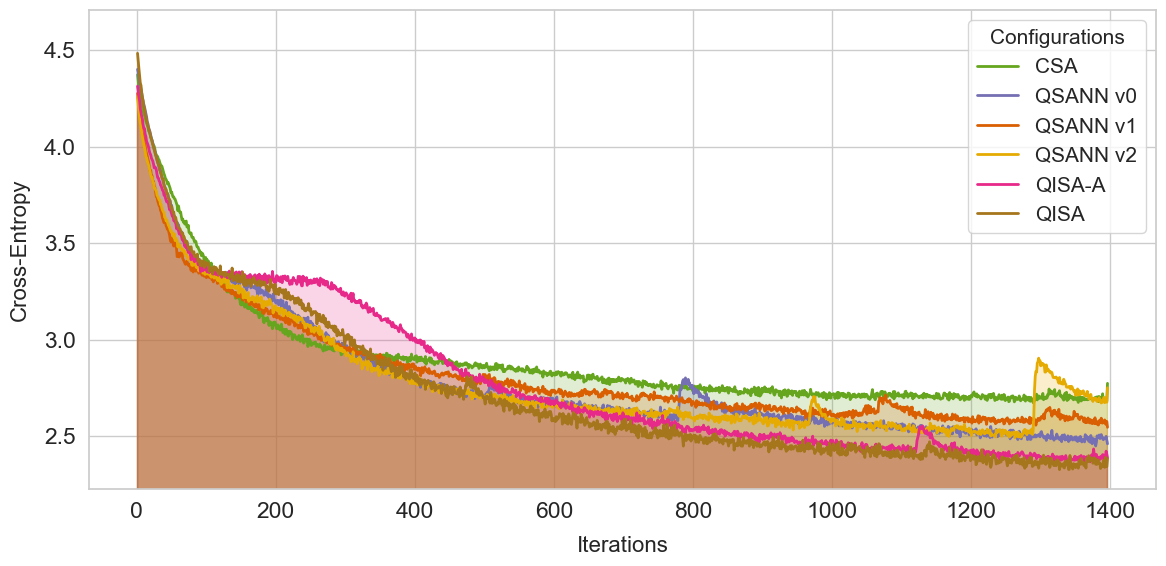}
    \end{subfigure}

    \vspace{5mm}

    \begin{subfigure}{\linewidth}
        \centering
        \textbf{Embedding size = 16, Number of heads = 1}\par\vspace{2mm}
        \includegraphics[width=1\linewidth]{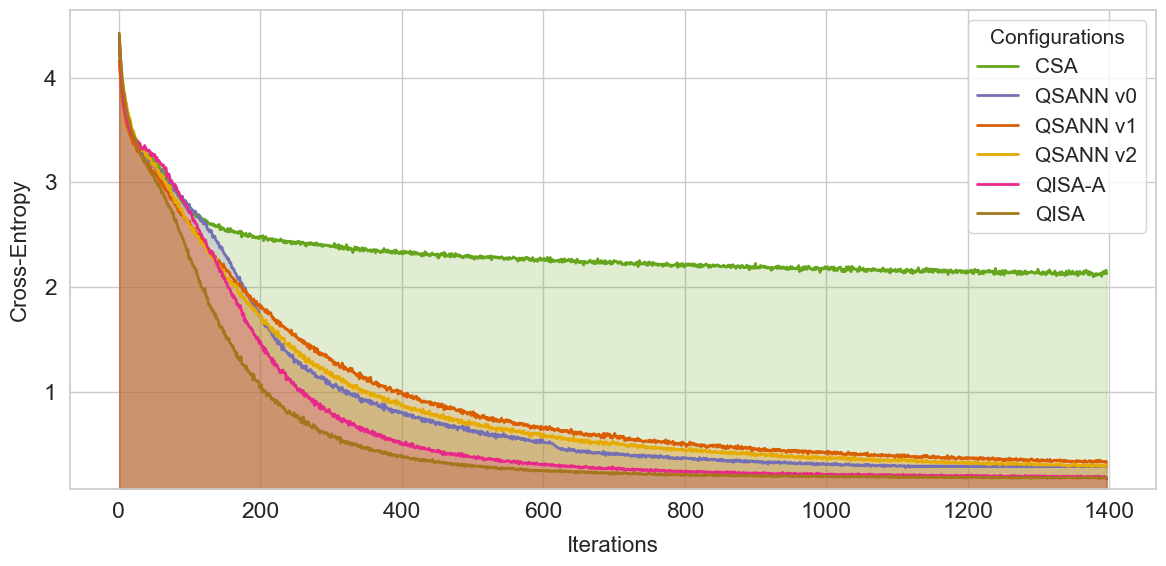}
    \end{subfigure}

    \vspace{5mm}

    \begin{subfigure}{\linewidth}
        \centering
        \textbf{Embedding size = 16, Number of heads = 4}\par\vspace{2mm}
        \includegraphics[width=1\linewidth]{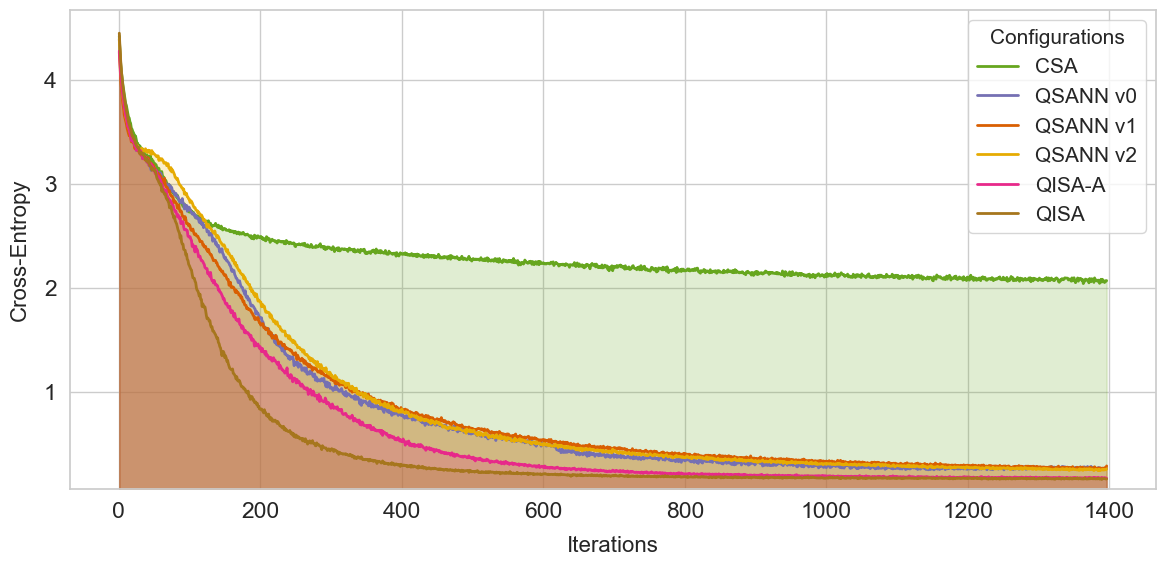}
    \end{subfigure}

    \caption{Cross-entropy versus iterations during the training of GPT-1 with CSA, QSANN (only for 1 head), QSANNv1, QSANNv2, and QISA. 
    The top plot is for embedding size 4 and 1 head, the middle is for embedding size 16 and 1 head, and the bottom is for embedding size 16 and 4 heads.}
    \label{fig:GPT1_loss_comparisons}
\end{figure}

Figure~\ref{fig:GPT1_loss_comparisons} illustrates the cross-entropy loss versus the number of iterations for all models considered. Here, training is performed during two epochs, which can be observed to be sufficient for convergence. The plots only depict results for quantum models using 3 layers of the HEA, since we found the behavior with 1, 2 and 3 layers to be almost identical. It can be observed that CSA obtains the highest cross-entropy loss of all models, and all quantum and quantum-inspired models obtain similar lower values. The cross-entropy gap between CSA and the other models widens for larger embedding sizes, with QISA achieving an improvement of $13\times$ over CSA when using an embedding size of 16.

In table Table \ref{tab:metrics} we compare the performance of the models under metrics used for natural language tasks: word error rate (WER) and character error rate (CER).
It can be observed that in these experiemnts, the quantum models do not benefit from the additional ansatz layers likely due to the shallow circuits being expressive enough for the modest embedding sizes used here. At embedding size 4, quantum and quantum-inspired models either outperform or match the CSA across all metrics, and notably QISA and QISA-A display a substantial advantage in WER. Mirroring the behavior observed in Figure \ref{fig:GPT1_loss_comparisons}, the performance gap widens when increasing the embedding size from 4 to 16. At embedding size 16, quantum and quantum-inspired models consistently outperform CSA, with QISA and QISA-A on the lead. It is also worth noting that we do not observe a significant performance difference between 1 and 4 heads at embedding size 16, likely also due to its modest embedding size. It is yet to be seen how the performance would change when using multiple heads for larger embedding sizes, and to what extent modern fully retrogressive models would benefit from using QISA, but this is beyond our current scope. 

%\onecolumngrid
\begin{table*}[t]
\centering
\begin{tabular}{|l|l|l|l|l|l|l|l|l|l|}
\hline
Model & \multicolumn{3}{|l|}{Embedding 4, 1 head} & \multicolumn{3}{l|}{Embedding 16, 1 head} & \multicolumn{3}{l|}{Embedding 16, 4 heads} \\
\cline{2-10}
& CE & CER & WER & CE & CER & WER & CE & CER & WER \\ \hline \hline
CSA
  & $2.72 \pm 0.05$ & $0.74 \pm 0.10$ & $1.34 \pm 0.48$
  & $2.16 \pm 0.07$ & $0.62 \pm 0.12$ & $1.17 \pm 0.36$
  & $2.16 \pm 0.07$ & $0.62 \pm 0.11$ & $1.18 \pm 0.37$ \\ \hline
QISA
  & $2.30 \pm 0.07$ & $0.68 \pm 0.11$ & $\mathbf{1.02 \pm 0.19}$
  & $\mathbf{0.16 \pm 0.00}$ & $\mathbf{0.04 \pm 0.02}$ & $\mathbf{0.25 \pm 0.19}$
  & $\mathbf{0.16 \pm 0.00}$ & $\mathbf{0.04 \pm 0.02}$ & $\mathbf{0.25 \pm 0.18}$ \\ \hline
\makecell[l]{QISA-A \\(1/2/3 layers)}
  & \makecell[l]{$\mathbf{2.27 \pm 0.06}$\\$2.28 \pm 0.07$\\$\mathbf{2.27 \pm 0.06}$}
  & \makecell[l]{$0.680 \pm 0.12$\\$0.679 \pm 0.10$\\$0.679 \pm 0.10$ }
  & \makecell[l]{$1.06 \pm 0.19$\\$1.05 \pm 0.20$\\$1.06 \pm 0.19$}
  & \makecell[l]{$\mathbf{0.16 \pm 0.00}$ \\ $0.17 \pm 0.01$ \\ $\mathbf{0.16 \pm 0.00}$}
  & \makecell[l]{$\mathbf{0.04 \pm 0.02}$ \\ $\mathbf{0.04 \pm 0.02}$ \\ $\mathbf{0.04 \pm 0.02}$}
  & \makecell[l]{$\mathbf{0.25 \pm 0.19}$ \\ $\mathbf{0.25 \pm 0.19}$ \\ $\mathbf{0.25 \pm 0.19}$}
  & \makecell[l]{$\mathbf{0.16 \pm 0.00}$ \\ $0.17 \pm 0.01$ \\ $\mathbf{0.16 \pm 0.00}$}
  & \makecell[l]{$\mathbf{0.04 \pm 0.02}$ \\ $\mathbf{0.04 \pm 0.02}$ \\ $\mathbf{0.04 \pm 0.02}$}
  & \makecell[l]{$\mathbf{0.25 \pm 0.19}$ \\ $\mathbf{0.25 \pm 0.19}$ \\ $\mathbf{0.25 \pm 0.19}$} \\ \hline
\makecell[l]{QSANN \\(1/2/3 layers)}
  & \makecell[l]{$2.34 \pm 0.07$ \\ $2.33 \pm 0.07$ \\ $2.35 \pm 0.08$}
  & \makecell[l]{$0.69 \pm 0.09$ \\ $0.68 \pm 0.09$ \\ $0.70 \pm 0.10$}
  & \makecell[l]{$1.34 \pm 0.45$ \\ $1.33 \pm 0.46$ \\ $1.35 \pm 0.45$}
  & \makecell[l]{$0.22 \pm 0.03$ \\ $0.21 \pm 0.03$ \\ $0.22 \pm 0.03$}
  & \makecell[l]{$0.06 \pm 0.04$ \\ $0.06 \pm 0.04$ \\ $0.07 \pm 0.04$}
  & \makecell[l]{$0.31 \pm 0.25$ \\ $0.30 \pm 0.25$ \\ $0.32 \pm 0.26$}
  & \makecell[l]{$0.19 \pm 0.01$ \\ $0.19 \pm 0.02$ \\ $0.20 \pm 0.00$}
  & \makecell[l]{$0.05 \pm 0.03$ \\ $0.05 \pm 0.04$ \\ $0.05 \pm 0.03$}
  & \makecell[l]{$0.28 \pm 0.21$ \\ $0.28 \pm 0.22$ \\ $0.28 \pm 0.22$} \\ \hline
\makecell[l]{QSANNv1 \\(1/2/3 layers)}
  & \makecell[l]{$2.35 \pm 0.07$ \\ $2.34 \pm 0.07$ \\ $2.36 \pm 0.07$}
  & \makecell[l]{$0.70 \pm 0.09$ \\ $0.71 \pm 0.09$ \\ $0.69 \pm 0.09$}
  & \makecell[l]{$1.33 \pm 0.46$ \\ $1.34 \pm 0.45$ \\ $1.32 \pm 0.47$}
  & \makecell[l]{$0.22 \pm 0.03$ \\ $0.23 \pm 0.03$ \\ $0.22 \pm 0.03$}
  & \makecell[l]{$0.06 \pm 0.04$ \\ $0.07 \pm 0.04$ \\ $0.06 \pm 0.04$}
  & \makecell[l]{$0.31 \pm 0.26$ \\ $0.32 \pm 0.25$ \\ $0.30 \pm 0.26$}
  & \makecell[l]{$0.19 \pm 0.01$ \\ $0.18 \pm 0.01$ \\ $0.18 \pm 0.01$}
  & \makecell[l]{$0.05 \pm 0.03$ \\ $0.05 \pm 0.03$ \\ $0.05 \pm 0.03$}
  & \makecell[l]{$0.27 \pm 0.21$ \\ $0.28 \pm 0.21$ \\ $0.26 \pm 0.20$} \\ \hline
\makecell[l]{QSANNv2 \\(1/2/3 layers)}
  & \makecell[l]{$2.29 \pm 0.07$ \\ $2.28 \pm 0.07$ \\ $2.30 \pm 0.07$}
  & \makecell[l]{$0.68 \pm 0.10$ \\ $\mathbf{0.67 \pm 0.10}$ \\ $0.69 \pm 0.10$}
  & \makecell[l]{$1.49 \pm 0.44$ \\ $1.50 \pm 0.44$ \\ $1.48 \pm 0.44$}
  & \makecell[l]{$0.22 \pm 0.02$ \\ $0.21 \pm 0.02$ \\ $0.22 \pm 0.02$}
  & \makecell[l]{$0.06 \pm 0.04$ \\ $0.06 \pm 0.04$ \\ $0.07 \pm 0.04$}
  & \makecell[l]{$0.30 \pm 0.24$ \\ $0.31 \pm 0.24$ \\ $0.29 \pm 0.24$}
  & \makecell[l]{$0.19 \pm 0.01$ \\ $0.17 \pm 0.00$ \\ $0.18 \pm 0.01$}
  & \makecell[l]{$0.05 \pm 0.03$ \\ $0.05 \pm 0.03$ \\ $0.05 \pm 0.03$}
  & \makecell[l]{$0.27 \pm 0.21$ \\ $0.26 \pm 0.20$ \\ $0.26 \pm 0.21$} \\ \hline
\end{tabular}
\caption{Comparison of cross-entropy loss (CE), character error rate (CER) and word error rate (WER) for self-attention models: standard (CSA), quantum-inspired (QISA), and quantum (QISA-A, QSANN, QSANNv1, QSANNv2) with different numbers of layers. Best results per metric are highlighted in bold. Setup: 2 pretraining epochs, batch size = 1024.}
\label{tab:metrics}
\end{table*}

% {\color{blue} We compare the performance of QISA and CSA trained for 2 epochs (denoted as 2e), and CSA for 30 epochs (30e). Tables~\ref{tab:metrics_emb4}--\ref{tab:metrics_emb16_4heads} show the results of these models for some performance metrics widely used for LLMs: character error rate (CER) and word error rate (WER). Under these metrics, QISA consistently achieves values compared to the CSA trained for 2 and 30 epochs, indicating that the better results come from model architectural improvements. }

 A comparison of the number of trainable parameters per head is presented in Table \ref{tab:full_parameter_comp}. The quantum models have an advantage in this respect, with QSANNv1 and v2 requiring the least number of trainable parameters. They are followed by QSANN which has a dependency on the context length, and QAISA-A which inherits $2mh$ classical parameters from CSA. The classical QISA and CSA require the highest number of parameters, with QISA requiring the most for $H>1$. QISA could benefit from techniques designed to reduce parameter counts e.g. low-rank factorization \cite{hu2021lora, fu2020computing, li2023losparse}, but it is beyond our current scope. Notice that for a single head $H=1$, QISA and CSA have the same parameter count but, as can be observed in Table \ref{tab:metrics}, for embedding sizes 16 and 4 with 1 head QISA remarkably outperforms CSA, indicating that the improvements are architectural rather than only related to parameter quantity.

\begin{figure}[h]
    \centering
    \includegraphics[width=1\linewidth]{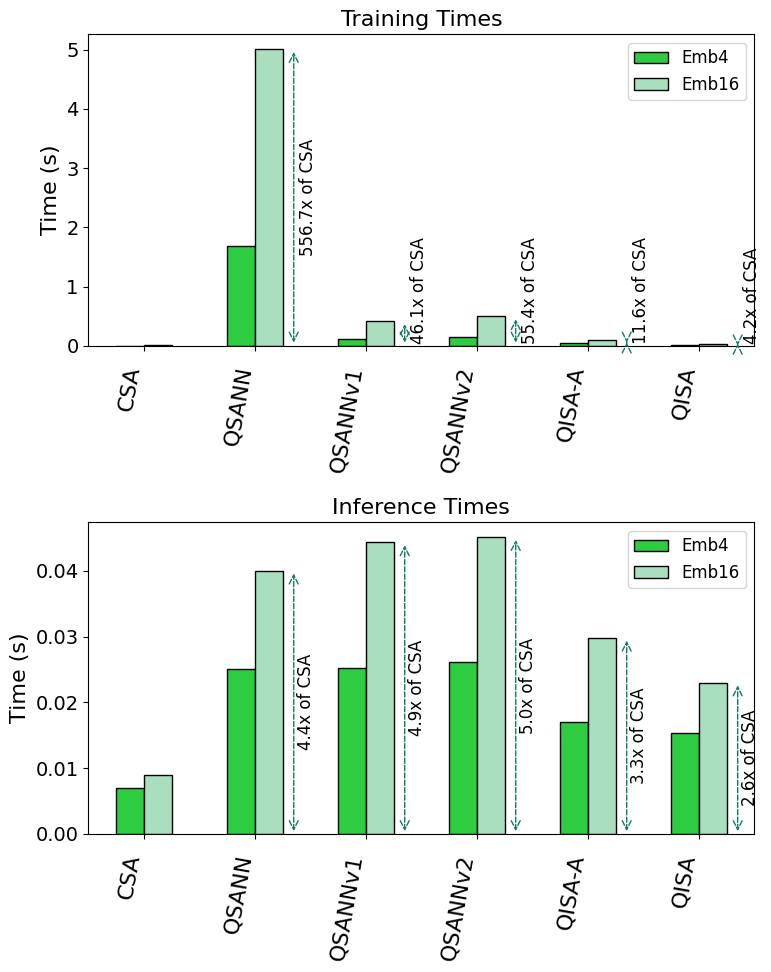}
        \caption{Training and inference times for a batch size of 1024 on a single NVIDIA T4 GPU. All models use observables caching to increase their inference speeds. }
    \label{fig:speed}
\end{figure}

Note that despite being fewer, training quantum parameters in simulation is more resource intensive than training classical parameters, as it requires the repeated calculation of the unitary matrix, resulting in a considerable overhead. This can be observed in Figure \ref{fig:speed}, which compares the training and inference times of the models studied. During training, the simulated quantum models require orders of magnitude more time compared to the classical models. During inference, since the unitaries are fixed, we cache the observables evolved in the Heisenberg picture to increase the inference speed of simulated quantum models (details in Appendix \ref{app:fast_inference}). A similar strategy is used to increase QISA's inference speed. Here, QISA has an inference time just $2.6\times$ slower than CSA. Given the advantage of QISA in performance metrics, the penalty of a slightly longer inference time may be a worthwhile trade-off.

Although the precise mechanism behind QISA’s superior performance metrics remains uncertain, its increased complexity appears to enhance the ability to perform more effective transformations. In particular, this is achieved while maintaining a reasonable number of parameters, which are reused through a process reminiscent of cross-layer parameter-sharing techniques \cite{lan2019albert, takase2023lessons, hu2024aslora}.

\begin{table}[H]
  \centering
  \small
  \setlength{\tabcolsep}{2pt}
  \renewcommand{\arraystretch}{1.1}
  \resizebox{0.6\columnwidth}{!}{%
    \begin{tabular}{|l|c|c|}
      \hline
      \textbf{Model} & \textbf{Parameters per head} \\ 
      \hline\hline
        QISA 
        & \(2m \times h + m^2\)  
      \\ 
      \hline
      CSA  
        & \(3m \times h\)  
       \\ 
      \hline
      QISA - A  
        & \(2m \times h + 3\,\lceil\log_2 m\rceil\times p\)  
      \\ 
      \hline
      QSANN 
        & \(3 \times 3\,\lceil\log_2 m\rceil \times p\times l\)  
       \\ 
      \hline
      QSANNv1  
        & \(3 \times 3\,\lceil\log_2 m\rceil\times p\)  
       \\ 
      \hline
      QSANNv2  
        & \(3 \times 3\,\lceil\log_2 m\rceil\times p\)  
       \\ 
      \hline
    \end{tabular}%
  }
  \caption{Number of parameters per head for CSA, QISA, QISA-A and QSANN models. Here $m$ is the embedding size, $h$ is the head output dimension ($h = m/H$ where $H$ is the number of heads), $l$ is the context size, the factor of 3 in quantum variants corresponds to the rotation parameters in each qubit ansatz layer,and $p$ is the number of ansatz layers.}
  \label{tab:full_parameter_comp}

\end{table}

% We provide all the training charts for CSA, QISA and improvements over QSANN in the Appendix section \ref{fig:GPT1_loss_comparisons} as well as tables comparing the number of parameters \ref{tab:full_parameter_comp}.

\section{Conclusions}

We propose two self-attention models:~a classical quantum-inspired self-attention mechanism (QISA) which swaps the value layer of the self-attention mechanism for operations inspired by quantum state evolution and expectation values; and a quantum deployable version (QISA-A), which requires fewer parameters and achieves a similar performance while being suited for future error-corrected quantum computers.
Both models take cues from the QSANN quantum self-attention mechanism \cite{li2024quantum} which has shown advantage over classical-self attention (CSA) in text classification tasks.

We implement CSA, QISA,  QISA-A and three variants of QSANN into the fully autoregressive model of GPT-1 \cite{vaswani2017attention}. To our knowledge, this is the first time such an implementation has been done for quantum self-attention mechanisms. Our numerical experiments show that QISA and QISA-A beat or match the other models in performance metrics. Compared to CSA using an embedding size of 16,  QISA and QISA-A obtain a $13\times$ improvement in cross-entropy loss, 15.5$\times$ in character error rate (CER), and 4.7$\times$ in word error rate (WER). This is at the expense of QISA having training and inference times that are respectively 4.2$\times$ and 2.6$\times$ longer than those of CSA, which may be considered a worthwhile trade-off.

In single head configurations, both QISA and CSA share the same number of parameters, still we find better results in performance metrics for QISA, indicating that the advantage stems from architectural improvements.
For configurations using more than a single attention head, the number of parameters of QISA exceeds that of CSA. This could be alleviated with the use techniques to reduce dimensionality. These possibilities will be investigated in future research.

Although the exact mechanisms underlying QISA’s superior performance metrics remain uncertain, the increased complexity of the quantum-inspired value layer appears to enable more effective transformations. Importantly, this added complexity does not lead to an excessive number of parameters, as the value layer leverages repeated application of the same linear map—a strategy reminiscent of cross-layer parameter sharing techniques \cite{lan2019albert, takase2023lessons, hu2024aslora}.

The absence of error correction in current quantum hardware prevents the reliable execution of quantum self-attention models. However, as quantum hardware advances, QISA-A could emerge as a viable alternative. If, for larger embedding sizes, QISA-A maintains performance comparable to QISA while requiring fewer parameters—thereby offsetting the computational cost of the parameter-shift rule—it may become the preferred approach in the future.

In conclusion, our quantum-inspired self-attention mechanism, QISA, inherits the advantages of both quantum and classical self-attention mechanisms: it matches or outperforms both quantum and classical models in multiple metrics; can be classically parallelized; and has training and inference speeds of the order of classical self-attention. 
Furthermore, our quantum deployable model, QISA-A, requires less parameters and has similar performance, making it a prime candidate for future error-corrected quantum computers.
We hope these results will lead to more performance improvements for LLMs and other natural language processing applications, and inspire the development of future models that incorporate elements of quantum computing.

\section*{Code availability}
The open-source repository written on the PyTorch + TorchQuantum frameworks can be found in \url{https://github.com/Nikait/QISA}.

\bibliography{refs.bib}
\bibliographystyle{unsrt}

\appendix

\section{QSANN variants}
\subsection{Original QSANN}
\label{App:QSANNv0}

The original QSANN self-attention mechanism introduced in \cite{li2024quantum} maps every token into an \(n\)-qubit quantum state via an encoding circuit:
\[
\ket{\psi_i} = U_{\mathrm{enc}}(x_i)\,\ket{0}^{\otimes n},
\quad i = 1,\dots,l,
\]
where \(l\) is the context length (number of tokens).

Then, three token-specific parameterized circuits
\(U_q^i(\theta_q^i)\), \(U_k^i(\theta_k^i)\), and \(U_v^i(\theta_v^i)\) are used to transform the encoding states. The resulting states are used to produce: a scalar ``query'' \(q_i\), a scalar ``key'' \(k_i\), and a vector ``value'' \(v_i\) as follows:
\[
\begin{aligned}
q_i &= \bra{\psi_i}U_q^{i\dagger}Z_1U_q^i\ket{\psi_i},\quad
k_i = \bra{\psi_i}U_k^{i\dagger}Z_1U_k^i\ket{\psi_i},\\
v_i &= \bigl(\bra{\psi_i}U_v^{i\dagger}P_jU_v^i\ket{\psi_i}\bigr)_{j=1}^m \in \mathbb{R}^m,
\end{aligned}
\]
where each \(P_j \in \{I,X,Y,Z\}^{\otimes n}\) is an \(n\)-qubit Pauli string.

Collecting these for all tokens \(i=1,\dots,l\) gives
\[
Q = (\,q_i\,)_{i=1}^l,\quad
K = (\,k_i\,)_{i=1}^l,\quad
V = (\,v_i\,)_{i=1}^l \;\in\;\mathbb{R}^{l\times m}.
\]

Instead of using a dot product between \(Q\) and \(K\), the attention weights are computed via a Gaussian kernel:
\[
A_{i,j} = \frac{\exp\!\left(- (q_i - k_j)^2 \right)}{\sum_{r=1}^l \exp\!\left(- (q_i - k_r)^2 \right)},
\]
so that \(A \in \mathbb{R}^{l\times l}\).

The block’s output is then
\[
Y = A\,V,
\]
an \(l\times m\) matrix of updated token embeddings.

\subsection{QSANNv1: The QSANN with reduced number of parameters}
\label{app:QSANNv1}

Here, the QSANN model is simplified by sharing $U_q$, $U_k$ and $U_v$ on all tokens, further reducing the number of parameters as shown in \ref{tab:full_parameter_comp}. In the metrics it performs similarly to the original QSANN, and thus better than CSA, details shown in Table \ref{tab:metrics}

\subsection{QSANNv2: The QSANN with same $Q/K$ structure as V}
\label{app:QSANNv2}

Further modified from QSANNv1, here the $Q$ and $K$ layers share a unified structure with the $V$ layer. This modification aims to improve the expressiveness of the attention matrix by going beyond the Gaussian kernel. Specifically, the query and key parts are constructed by the expectation values of observables $P_j \in \{I, X, Y, Z\}^{\otimes n}$, rather than measuring a single qubit measurement. The rows of $Q$ and K are computed as: 
% \todo[inline]{EC: can you confirm it is indeed rows?}
% \todo[inline]{NK: Yes, but the notation is slightly different than in the description of the original version, there were no square brackets}
\begin{align}\label{eq:qisav2_qk}
q_i &= \bigl[\langle P_1\rangle_q^i,\;\langle P_2\rangle_q^i,\;\dots,\;\langle P_m\rangle_q^i\bigr]\\
k_i &= \bigl[\langle P_1\rangle_k^i,\;\langle P_2\rangle_k^i,\;\dots,\;\langle P_m\rangle_k^i\bigr],
\end{align} 
where $\langle P_j\rangle_q^i = \langle {\psi_i} | U_q^{\dagger}(\theta_q) P_j  U_q(\theta_q) | {\psi_i} \rangle$ for $Q$, and similarly for $K$, with shared parameters for all tokens. As shown in Table \ref{tab:metrics}, this model performs similarly to the other QSANN variants, indicating the performance gains come primarily from the $V$ layer structure.

% \subsection{Does the Depth of HEA Matter?}

% To investigate the influence of the hardware-efficient ansatz (HEA) depth in the QISA model, we conducted experiments with 1, 2, and 3 HEA layers in the Value circuit while keeping other parameters fixed (embedding size = 16, 1 head, 2 pretraining epochs, batch size 1024). 

% The results, summarized in Table~\ref{tab:hea_depth} in Appendix~\ref{app:evaluation_metrics}, show that increasing the HEA depth does not lead to improvements in cross-entropy loss or other evaluation metrics on the test set. However, deeper HEA circuits significantly increase training time per batch, while inference time remains unchanged.

% The role of quantum parameters in the single HEA layer of QISA was also investigated. To understand what drives the strong performance of QISA, we conducted an ablation where all parameterized quantum operators were removed, leaving only the amplitude encoding and the original measurement scheme relative to the Pauli operators intact. A classical linear layer projecting from \( m \) to \( m \) was added to replace the parametrized quantum operations. Interestingly, this modified setup yielded similar evaluation metrics, suggesting that the nonlinearities induced by the quantum measurements themselves—rather than the trainable quantum parameters—are the primary contributors to the improved performance of QISA (see Table~\ref{tab:qisa_frozen_params}).

% \todo[inline]{move the tables comparing QISA and QISA-A here, and the layers effect}

\section{Fast Inference}
\label{app:fast_inference}

During training we simulate the quantum circuits and compute gradients via reverse-mode automatic differentiation (i.e., backpropagation through the quantum circuit), a process that requires the repeated computation of the unitary matrices. However, at inference - when no gradient computation is needed - we precompute and cache these observables and evolve in the Heisenberg picture to achieve significantly faster forward passes.

We index by transformer layer $\ell$ and attention head $j$.  After training, every variational parameter is fixed.  Then, for each $(\ell,j)$, the unitary block   
\[
U_{\mathrm{total}}^{(\ell,j)}
\;=\;
\prod_{g=1}^{G}
U_g^{(\ell, j)},
\]
where $U_g^{(\ell, j)}$ are the individual unitary gates, and $G$ is the number of gates in the ansatz.
These are used to evolve each Pauli string $P_k$ in the Heisenberg picture as 
\[
P_k'^{(\ell, j)}
\;=\;
U_{\mathrm{total}}^{(\ell, j) \dagger}\;P_k\;U_{\mathrm{total}}^{(\ell,j)},
\]
and store each $P_k'^{(\ell, j)}$ as $2^n \times 2^n$ matrices.

At inference time, the expectation of $P_k$ for $(\ell,j)$ with respect to an input token $\ket{x_i}$ is given by  the inner product:
\[
\langle P_k\rangle_i^{(\ell, j)} = \bra{x_i} P_k'^{(\ell, j)}\ket{x_i}
\]

requiring only one matrix–vector and one vector–vector multiplication per observable.

Similarly, for QISA we precompute the operators 

\[
P_k'^{(\ell, j)}
\;=
\tilde{W}_V^{(l,j)\intercal}P_k\;\tilde{W}_V^{(l,j)},
\]
and store each $P_k'^{(\ell, j)}$ as $2^n \times 2^n$ matrices.
% In QISA, this observable precomputation halves the remaining inference overhead, so that combined with unitary precomputation the model runs only at $\approx3\times$ the cost of classical self‑attention.  
% The procedure performed in a classical simulator and has a computational complexity of $O(lm^3)$. Additionaly we find the computation of $\tilde{V}$ to be more efficient than $V$ when $l>m^2+m$.

% \todo[]{NK: I've updated the plot, made the font as large as possible:(. Also updated the caption regarding to precomputation of single version.}

\end{document}